\def\year{2015}
\documentclass[letterpaper]{article}
\usepackage{aaai16}
\usepackage{times}
\usepackage{helvet}
\usepackage{courier}
\usepackage{amsmath}
\usepackage{graphicx}
\usepackage[utf8]{inputenc}
\usepackage[normalem]{ulem} 
\usepackage{amssymb}
\usepackage{enumitem}

\begin{document}
%
 \title{MUST-CNN: A Multilayer Shift-and-Stitch Deep Convolutional Architecture for Sequence-based Protein Structure Prediction}
\author{
Zeming Lin \\
Department of Computer Science\\
University of Virginia\\
Charlottesville, VA 22904 \\
\texttt{zl4ry@virginia.edu}
\And
Jack Lanchantin \\
Department of Computer Science\\
University of Virginia\\
Charlottesville, VA 22904 \\
\texttt{jjl5sw@virginia.edu }
\And
Yanjun Qi \\
Assistant Professor \\ 
Department of Computer Science\\
University of Virginia\\
Charlottesville, VA 22904 \\
\texttt{yq2h@virginia.edu }
}

\maketitle
\begin{abstract}
\begin{quote}
    Predicting protein properties such as solvent accessibility and secondary structure from its primary amino acid
    sequence is an important task in bioinformatics.
    Recently, a few deep learning models have surpassed the traditional window based multilayer perceptron.
    Taking inspiration from the image
    classification domain we propose a deep convolutional neural network architecture, MUST-CNN, to predict
    protein properties.
    This architecture uses a novel multilayer shift-and-stitch (MUST) technique to generate fully dense per-position predictions on
    protein sequences. Our model is significantly simpler than the state-of-the-art, yet achieves better results.
    By combining MUST and the efficient convolution operation, we can consider far more parameters
    while retaining very fast prediction speeds. We beat the state-of-the-art performance on
    two large protein property prediction datasets.
\end{quote}
\end{abstract}

\section{Introduction}
Proteins are vital to the function of living beings. It is easy to determine the sequence of a protein, yet it is
difficult to determine other properties, such as secondary structure and solvent accessibility.
These properties are hypothesized to be almost uniquely determined by primary structure, but it is still computationally
difficult to determine them on a large scale.

Previous state-of-the-art methods for protein secondary structure prediction use multilayer perceptron (MLP) networks
\cite{qi_unified_2012,drozdetskiy_jpred4:_2015}. In order to predict a per-position label for each amino acid in the
input protein sequence, MLP networks must use a ``windowing'' approach where a single label is predicted by feeding the target 
amino acid and its surrounding amino acids through the network. This is then repeated for each amino acid in the sequence. 
These architectures generally has two major drawbacks due to the windowing approach: (1) they take a long time to train, on 
the order of days or weeks and (2) they have smaller window sizes, and thus cannot make longer range connections.
For instance, the PSIPRED algorithm handles a window size of only 15 amino acids \cite{jones_protein_1999}.

To overcome the windowing issue, we propose to use a convolutional neural network (CNN) which can label the properties of
each amino acid in the entire target sequence all at once. CNNs have been used successfully in computer vision
\cite{pinheiro_recurrent_2013,szegedy_going_2014}
and natural language processing \cite{kim_convolutional_2014,collobert_unified_2008}.
In addition to parameter sharing and pooling, which reduce computation, CNNs are also highly parallelizable. 
Thus, CNNs can achieve a much greater speedup compared to a windowed MLP approach. The issue when trying to label each
position in an input sequence with a CNN is that pooling leads to a decreased output resolution.
To handle this issue, we propose a new
multilayer shift-and-stitch method which allows us to efficiently label each target input at full resolution in a 
computationally efficient manner.

We show that a MUltilayer Shift-and-sTitch CNN (MUST-CNN) trained end-to-end and per-position on
protein property prediction beats the state-of-the-art without other machinery.
To our knowledge, we are the first to train convolutional networks end-to-end for per-position
protein property prediction. Both learning and inference are performed on entire arbitrarily sized
sequences. Feedforward computation and backpropagation is made possible by our novel application of
the shift-and-stitch technique on the entire sequence.

In summary we make the following contributions:
\begin{enumerate}
    \item   Beat the state-of-the-art performance on two large datasets of protein property prediction tasks.
    \item   Propose a multilayer shift-and-stitch technique for deep CNNs,
            which significantly speeds up training and test time and
            increases the size of the model we can train.
    \item   Propose a generic end-to-end system for per-position labeling on the sequence level. That is, for a sequence
            $\{a_k\}_{k=1}^n$, we can generate labels $\{y_k\}_{k=1}^n$ for each $a_k$.
\end{enumerate}

\section{Related Works}
Two of the most used algorithms in bioinformatics literature for protein property prediction are PSIPRED \cite{jones_protein_1999} 
and Jpred \cite{drozdetskiy_jpred4:_2015}. PSIPRED 3.2, which uses a two layer MLP approach,
claims a 3-class per-position accuracy (Q$_3$)
score of 81.6\%. The Jpred algorithm uses a
very similar structure of a two layer MLP network. However, Jpred considers more features and uses a jury based approach
with multiple models \cite{cuff_application_2000}. Jpred claims an 81.5\% Q$_3$ score on secondary structure
prediction, and also predicts relative solvent accessibility.
\cite{qi_unified_2012} uses a deep MLP architecture with multitask learning and achieves 81.7\% Q$_3$.
\cite{zhou_deep_2014} created a generative
stochastic network to predict secondary structure from the same data we used, for a Q$_8$ of 66.4\%. 
Unlike Q$_3$, the Q$_8$ accuracy tries to distinguish between more classes.

The state-of-the-art protein sequence classification system is SSpro, which obtains 91.74\% Q$_3$ and 85.88\% Q$_8$ on
a different unfiltered PDB dataset \cite{magnan_sspro/accpro_2014}. However, this system exploits
additional information via sequence similarity, and their reported accuracies were only
80\% without this module. Our work would complement their machine learning module and
likely result in even better accuracies.

Recently, work has also been done on the model side, particularly in natural language processing and image recognition tasks. \cite{collobert_natural_2011} created a similar algorithm in the natural language
processing domain, where they labeled word properties, such as part of speech or category of a named entity, on text
data. If we consider each protein chain to be a sentence and each amino acid to be a word, the techniques
transfer easily. \cite{collobert_natural_2011} used both a windowed approach and a sentence level
approach with a convolutional network, though their network was shallow and only outputed predictions for
one position at a time.
Long-short term memory networks have been used very successfully in sequence learning, machine translation
\cite{sutskever_sequence_2014,bahdanau_neural_2014} and language modeling \cite{zaremba_recurrent_2014}. We note
that machine translation is a much more general sequence to sequence task where the input and output sizes are
not matched. Language modeling tries to guess future words based on past words, while protein
sequences has no innate direction.

In the image domain, \cite{szegedy_going_2014} has beaten the state-of-the-art on image classification
by a large percentage through using a deep multilayer convolutional network in the ImageNet Large-Scale Visual
Recognition Challenge. Scene labeling is the task of labeling
each pixel of an image with one of several classes, a 2D analogue of protein property prediction.
\cite{pinheiro_recurrent_2013} uses a recurrent neural network to obtain state-of-the-art results on scene 
labeling without any feature engineering. \cite{long_fully_2014} designs fully convolutional networks for
dense pixel prediction by running several convolutional networks on different scales. \cite{sermanet_overfeat:_2013}
increases the resolution of a bounding box based image classifier by introducing the shift-and-stitch technique,
which we use on sequences instead of images and on the entire model instead of only on the last layer.

\section{Method: MUST-CNN}
\subsection{Convolutional Neural Networks (CNN)}

\begin{figure}
\begin{center}
    \includegraphics[width=\columnwidth]{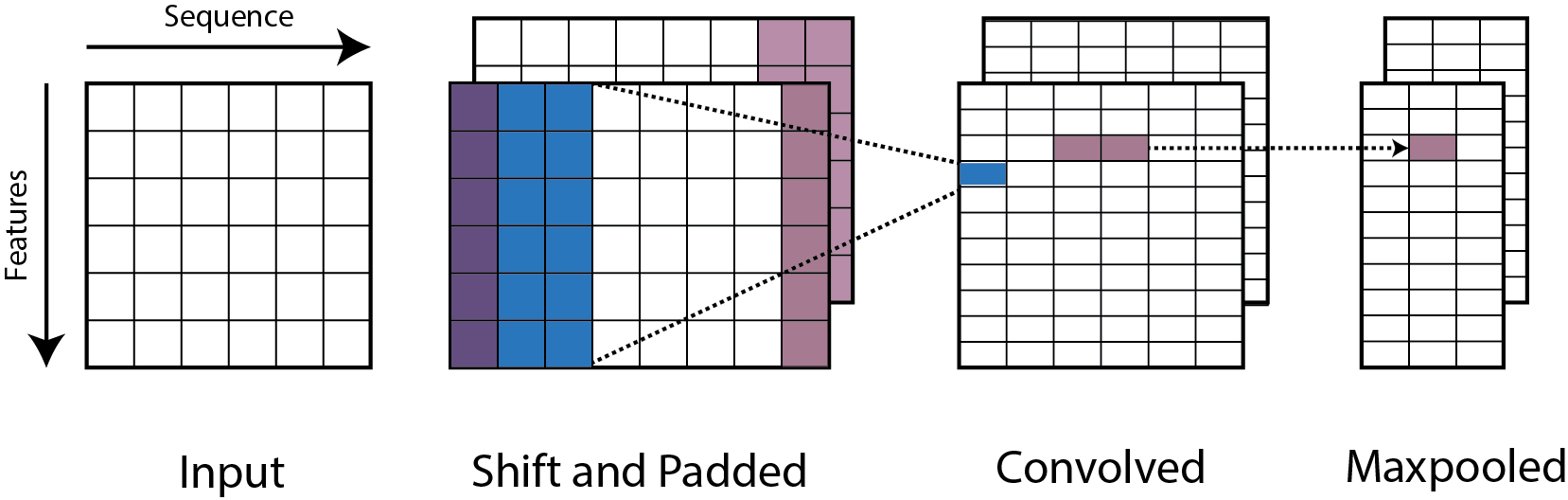}
\end{center}
\caption{
    A diagram for one layer of the convolutional network. We shift and pad the input accordingly
    to be able to label every position in our neural network. How the output is recombined into a fully dense
    per-position sequence label is given in Figure \ref{fig:maxpool}.
} \label{fig:layer}
\end{figure}

Convolutional networks were popularized for the task of handwriting recognition of 2D images \cite{lecun_gradient-based_1998}. 
In a similar way, we use a 1D convolution for the protein sequence labeling problem. A convolution
on sequential data tensor $\mathbf{X}$ of size $T\times n_{in}$ with length $T$, kernel size $k$ and input hidden
layer size $n_{in}$ has output $\mathbf{Y}$ of size $T\times n_{out}$:

$$\mathbf{Y}_{t,i} = \sigma(\mathbf{B}_i + \sum_{j=1}^{n_{in}} \sum_{z=1}^{k} \mathbf{W}_{i,j,k} \mathbf{X}_{t+z-1,j})$$

where $\mathbf{W}$ and $\mathbf{B}$ are the trainable parameters of the convolution kernel, and $\sigma$ is the
nonlinearity. We try three different nonlinearity functions in our experiments: the hyperbolic tangent,
rectified linear units (ReLU), and piecewise rectified linear units (PReLU). The hyperbolic tangent
is historically the most used in neural networks, since it has nice computational properties that
make optimization easy. Both ReLU and PReLU have been shown to work very
well on deep convolutional networks for object recognition. 
ReLU was shown to perform better than tanh on
the same tasks, and enforces small amounts of sparsity in neural networks \cite{glorot_deep_2011}.
By making the activations trainable and piecewise, PReLUs have
shown to match the state of the art on ILSVRC while converging in only 7\% of the time \cite{he_delving_2015}.

The ReLU nonlinearity is defined as

$$\mathrm{relu}(x) = \max(0, x)$$

and the PReLU nonlinearity is defined as

$$\mathrm{prelu}(x) =
\begin{cases} 
    \alpha x & \mbox{if } x < 0 \\
    x        & \mbox{if } x \ge 0
\end{cases}$$

with a trainable parameter $\alpha$.

After the convolution and nonlinearity, we use a pooling layer.
The only pooling strategy tested was maxpooling, which has shown to perform much better than
subsampling as a pooling scheme \cite{scherer_evaluation_2010} and has generally been the preferred pooling strategy
for large scale computer vision tasks. Maxpooling on a sequence $\mathbf{Y}$ of size $T\times n$ with a pooling 
size of $m$ results in output $\mathbf{Z}$ where

$$\mathbf{Z}_{t,i} = \max_{j=1}^m { \mathbf{Y}_{m(t-1)+j, i} }$$

Finally, the outputs are passed through a dropout layer. The dropout layer is a randomized mask of the outputs,
equivalent to randomly zeroing out the inputs to the next layer during training time with probability $d$
\cite{srivastava_dropout:_2014}. During testing, the dropout layer is removed and all weights are used. 
This acts as a regularizer for the neural network and prevents overfitting, though the best values for $d$ 
must be discovered experimentally.

One layer of the convolutional network is depicted in Figure \ref{fig:layer}. In our model design, we apply the
CNN module multiple times for a deep multilayer framework.

\subsection{Multilayer Shift-and-Stitch (MUST)}

Pooling is a dimension reduction operation which takes several nearby
values and combines them into a single value -- maxpooling uses the max function to do this.
Maxpooling is important because as nearby values are merged into one,
the classifier is encouraged to learn translation invariance.
However, after a single application of maxpooling with a pool size of $m$ on input sequence $\mathbf{X}$ of length $T$, 
the resulting maxpool output has sequence length $\frac{T}{m}$.

Since the dimensionality of the sequence has been
divided by a factor of $m$, it is no longer possible to label every position of the original sequence. 
A technique to increase the resolution in convolutional networks was given in
\cite{sermanet_overfeat:_2013}, called ``shift-and-stitch''. Their implementation uses the technique in a 
two dimensional setting to increase the resolution of pixel labels in the last layer of a convolutional network,
for up to a $4\times$ increase in resolution. We observe that the limiting factor on applying this to an entire
image is the massive slowdown in computation, since each pooling layer in a two-dimensional case 
requires the network to stitch together 4 different outputs and 3 pooling layers require 64 different stitched
outputs. 

\begin{figure*}
\begin{center}
    \includegraphics[width=\textwidth]{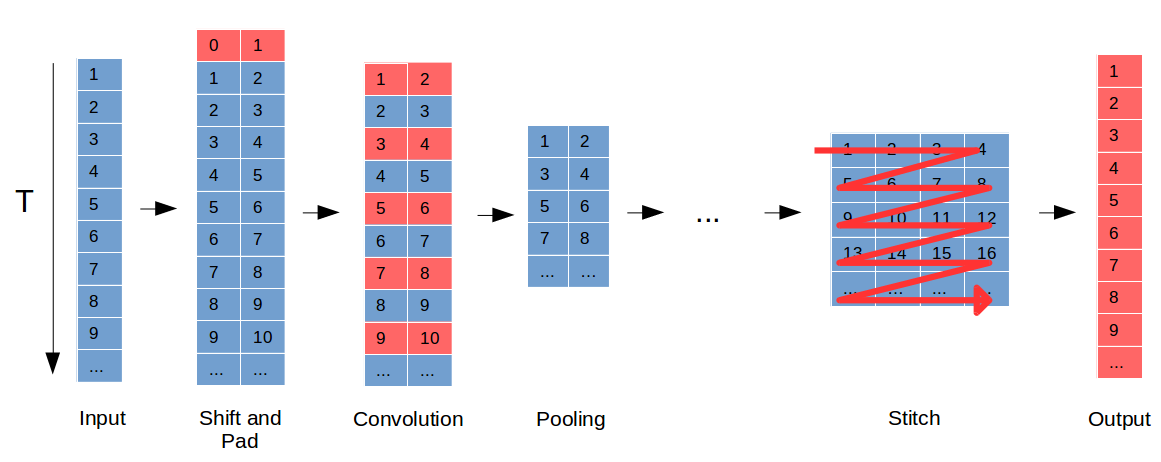}
\end{center}
\caption{
    Shift-and-stitch allows us to tag every element of an input even though maxpooling downsamples inputs.
    By zero padding each sequence correctly, we can join them along the batch dimension and process
    different shifts at the same time. This technique generalizes arbitrarily to any number of layers, and
    we can stitch together the result by rearranging and reshaping the resultant tensor, making
    computation very efficient.
}
    \label{fig:maxpool}
\end{figure*}

However, in the sequential case, we need to stitch together
significantly fewer sequences. Using 3 pooling layers with pooling size 2 will only requires 8 different stitches,
making computation tractable. Therefore, we propose to apply shift-and-stitch to every layer of our deep CNN which
generates dense per-position predictions for the entire sequence. This process is described in Figure \ref{fig:maxpool}.
This will allow us to take advantage of the computational speeds provided by the convolution module, making it
feasible to try a much larger model.

Due to the kernel size, a convolution with kernel size $k$ removes the $\lfloor\frac{k}{2}\rfloor$ edge values on 
each end of the sequence. Thus, we pad the input with a total of $\lfloor\frac{k}{2}\rfloor-1$ zeros at each end, colored as red in Figures \ref{fig:layer} and \ref{fig:maxpool}.
Because a maxpooling operation with pooling size $m$ labels every $m$ values in the input, we duplicate
the input $m$ times and pad the $i$-th input such that the first convolution window is centered on the first
amino acid. We observe that we can then join the $m$ duplicated inputs along the batch dimension and pass
it into the convolution module and take advantage of the batch computation ability offered by standard linear
algebra packages to train our system even faster. After pooling, the output is a zipped version of the original
input along the batch dimension. We simply ``stitch'' together the output in full resolution for the final result.

This novel multilayer shift-and-stitch technique makes it feasible to train a CNN end-to-end and generate dense per-position
protein property prediction.
This technique allows us to use convolution and maxpooling layers to label sequences of arbitrary length.

MUST can also be extended to train sequences in minibatches if needed, though the operations will be slightly
more complicated. However, we found minibatches not useful, because each amino acid is a training example, and each sequence
already contains many amino acids. Additionally, sequences are generally of different lengths, which make
implementation of minibatches harder.

\subsection{End-to-end Architecture}

\begin{figure}
    \begin{center}
        \includegraphics[width=\columnwidth]{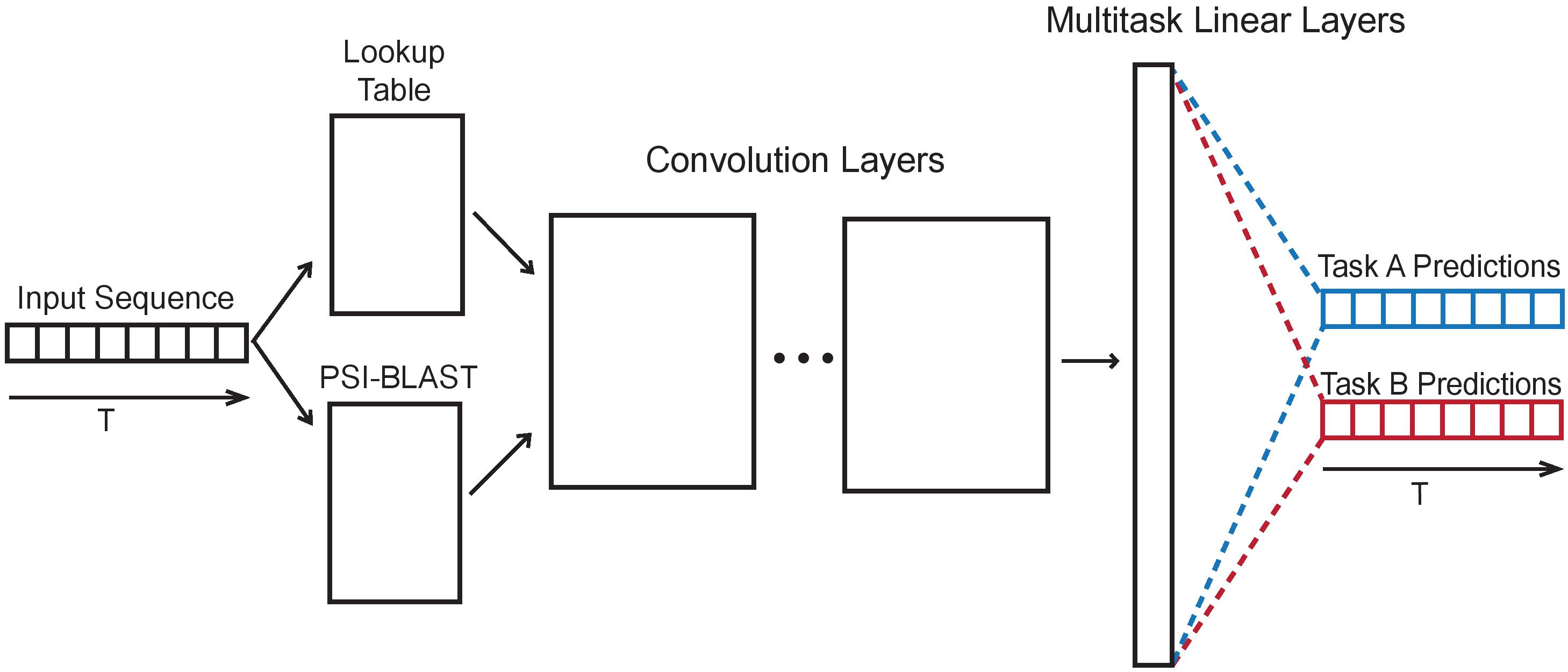}
    \end{center}
    \caption{
        An overview of the deep architecture of our model. Our model accepts an input protein sequence of length $T$,
        which is fed through the network to generate per-position predictions of length $T$ for several tasks.
    }
    \label{fig:network}
\end{figure}

In this section we describe the end-to-end model structure of the MUST-CNN and how we are able to train it to
make fully dense per-position predictions.

The input into the network is a one-hot encoding of an amino acid base pair sequence and the PSI-BLAST position
specific scoring matrix (PSSM), which is described in more detail in section Experiments subsection Feature.
Dropout is applied to the amino acid input and then fed through
a Lookup Table, similar to \cite{collobert_natural_2011}, to construct an embedding representation for each amino acid.
Then, the features from the amino acid embeddings are joined directly with the PSSM matricies along the feature dimension and fed into the deep convolutional network.

To apply the shift-and-stitch technique, we shift the amino acid sequences according to the amount of
pooling in each layer. Then, we pass every shift through each layer as described above, and stitch the
results together after all convolutional layers. This creates a deep embedding for every amino acid in our sequence.
Most previous methods use windowing to label the center amino acid. In our model, we can run the whole sequence
through the model instead of each window at a time.
This allows us to take advantage of the speed of convolution operations and use much larger
models.

We use a multitask construction similar to \cite{qi_unified_2012}, where we pass the deep 
embedding from the convolution layers into several linear fully connected layers which classify the protein sequence into each
separate task. This assumes a linear relationship between the deep embedding of a protein chain and
the properties predicted. In order for us to classify the outputs of the network for task
$\tau\in\mathcal{T}$, into class $c\in C_\tau$ for sequence $s\in S$, we apply the softmax
operator on the outputs $f_{t,\tau,c,s}$ of the subclassifiers for task $\tau$ at position $t=1,\ldots,T$.
Given the parameters of the network $\theta$, this gives us a conditional probability of class $c$:

$$p_\tau(c\in C_\tau|f_{t,\tau,s},\theta) = \frac{e^{f_{t,\tau,c,s}}}{\sum_{c \in C_\tau} e^{f_{t,\tau,c,s}}}$$

The parameters of the network are trained end-to-end by minimizing the negative log-likelihood function over the
training set, summing over all tasks and all elements in the sequence:

$$L(\theta) = -\sum_{s\in S}\sum_{\tau\in\mathcal{T}}\sum_{t=1}^T {\ln p_\tau(c_{correct} | f_{t,\tau,s}, \theta)}$$

where $c_{correct}$ is the correct label of the amino acid. The minimization of the loss function is obtained
via the stochastic gradient descent (SGD) algorithm with momentum, where we update the parameters after every sequence.
After the initial multitask model is trained, we take
the top layers and each task-specific subclassifier and \emph{fine-tune} the models by initializing their
weights at the weights learned by the multitask model and training only on each specific task with $\frac{1}{10}$
of the original learning rate. Regularization is achieved via dropout \cite{srivastava_dropout:_2014}.

All models are implemented using the Torch7 framework \cite{collobert_torch7:_2011}.

\subsection{Connecting to Previous Studies}

MUST-CNN is closely related to three previous models: OverFeat \cite{sermanet_overfeat:_2013}, Generative
Stochastic networks (GSNs) \cite{zhou_deep_2014}, and Conditional Neural Fields (CNFs) \cite{wang_protein_2011}.

CNFs are equivalent to a Conditional Random Field (CRF) with a convolutional feature extractor.
As far as we know, the authors implement a
windowed version using MLP networks. Their model, although able to consider the entire sequence due to the use of
a CRF, is unable to build deeper representations of models. Our model uses multiple convolutional layers and multitasking
to classify each amino acid into one of a few classes across multiple tasks. Our models are much deeper, and hence
can learn more efficient representations for complex dependencies.

The GSN is similar to a Restricted Boltzmann Machine with interconnections between the hidden states.
Training requires a settling algorithm
similar to finding the stationary distribution of a Markov chain. Although this technique allows for a model that considers
the entire protein sequence, it is less well understood. Convolution layers have the advantage of being used
more often in industry (See Related Works), and being well understood. Additionally, a fully feedforward
model is almost certainly faster than a model that requires a distribution to converge, though \cite{zhou_deep_2014}
did not state training or testing time in their paper.

OverFeat is the most closely related, though it works on images instead of sequence based classification.
The pipeline of OverFeat takes in images and classifies them densely to detect objects at every patch. Then
the bounding boxes for the objects are combined into a single bounding box, which is used to localize
the object. MUST-CNN is a one dimensional classification algorithm, which takes in the protein sequence surrounding
an amino acid and returns a dense property prediction of each amino acid. However, since object localization
does not need to be done on every bounding box, OverFeat only uses shift-and-stitch on the last layer for a small
resolution improvement. We do fully end-to-end shift-and-stitch, which is difficult on the image domain due to 
the quadratic increase in calculation time.

\section{Experiments}
\subsection{Feature}
The features that we use are (1) individual amino acids and (2) PSI-BLAST information \cite{altschul_gapped_1997} 
of a protein sequence. Each amino acid $a$ $\in$ $A$, where $A$ is the dictionary of amino acids, 
is coded as a one-hot vector in $\mathbb{R}^{\mid A \mid}$. That is, the encoding $x$ of the $i$-th amino acid
has $x_i=1$ and $x_{j\ne i}=0$.
PSI-BLAST generates a PSSM of 
size $T\times20$ for a $T$ lengthed sequence, where a higher score represents a higher likelihood 
of the $i$\textsuperscript{th} amino acid replacing the current one in other species. Generally, 
two amino acids that are interchangeable in the PSSM indicates that they are also interchangeable 
in the protein without significantly modifying the functionality of the protein.
The PSI-BLAST profiles were generated in the same way as the original authors in each of the datasets
\cite{qi_unified_2012,zhou_deep_2014}.

\subsection{Data}
We used two large protein property datasets in our experiments. The train, validation and test splits are
given in Table \ref{table:stats}. The two datasets we use are as follows:

\begin{description}[font=\normalfont\itshape]
    \item[4prot] Derived from \cite{qi_unified_2012}, we use a train-validation-test split where the
        model is trained on the training set, selected via validation set results, and best results reported
        by testing on the test set.
    \item[CullPDB] Derived from \cite{zhou_deep_2014}, we choose to use the CullPDB dataset where
        sequences with $>25\%$ identity with the CB513 dataset was removed. The train and validation sets are derived
        from CullPDB while the test set is CB513 in order to compare results with
        \cite{kaae_sonderby_protein_2014,wang_protein_2011}.
\end{description}

\subsection{Tasks}
Both datasets were formatted to have the same multitask representation. 
These are the four classification tasks we tested our system on: 

\begin{description}[font=\normalfont\itshape]
    \item[dssp]  The 8 class secondary structure prediction task from the dssp database \cite{touw_series_2015}.
        The class labels are H = alpha helix, B = residue in isolated beta bridge, E = 
        extended strand, G = 3-helix, I = 5-helix, T = hydrogen bonded turn, S = bend, L = loop.
    \item[ssp]   A collapsed version of the 8 class prediction task, since many protein secondary
        structure prediction algorithms use a 3 class approach instead of the 8-class approach
        given in dssp. $\{H,G\}\rightarrow H = $Helix, $\{B,E\}\rightarrow B= $Beta sheet, and
        $\{I,S,T,L\}\rightarrow C= $Coil
    \item[sar]   Relative solvent accessibility. Given the most solvent accessible amino acid in the protein
        has $x$ {\AA} of accessible surface area, we label other amino acids as solvent
        accessible if they have greater than $0.15x$ {\AA} of accessible surface area.
    \item[saa]   Absolute solvent accessibility. Defined as the amino acid having more than 0.15 {\AA} of
        accessible surface area.
\end{description}


\begin{table}
    \begin{center}
        \begin{tabular}{r | r | c c c}
           Datasets &   Number of      &   train   &   validation  &   test \\\hline
           4prot    &   Protein chains  &   7076    &   2359        &   2359 \\
                    &   Amino Acids     &   1500k   &   509k        &   506k \\\hline
            \small{CullPDB}  &   Protein chains  &   4427    &   1107        &   513 \\
            \small{\& CB513}&   Amino Acids     &   949k    &   235k        &   85k \\
        \end{tabular}
    \end{center}
    \caption{
        Size of datasets. We do a 60-20-20 split between training, test, and validation datasets on 4prot,
        but a 80-20-0 split on CullPDB, since we are testing on CB513.
    }
    \label{table:stats}
\end{table}
\subsection{Training}


\begin{table}
    \centering
    \begin{tabular}{r | l | l }
                          &   MUST-CNN    & MUST-CNN    \\
                          &   small       & large    \\\hline
        Convolution Layers&   3           & 3             \\
        Hidden units      &   189         & 1024          \\  
        Convolution Size  &   9           & 5             \\
        Maxpooling Size   &   2           & 2             \\
        Input Dropout     &   .35         & .1            \\
        Dropout           &   0           & \{.5, .3\}           \\
        Nonlinearity      &   ReLU        & ReLU           \\
        Learning Rate     &   0.0148      & 0.01        \\
        Momentum          &   0.9         & 0.9         \\
    \end{tabular}
    \caption{
        Model parameters for all models. The parameters on the small model were discovered via Bayesian
        Optimization, while the parameters on the large model were discovered using grid search assisted manual
        tuning. The dropout on the large network was 0.5 on CullPDB, but 0.3 on 4prot, adjusted based on the
        difference between training and validation error. All models were trained for 50 iterations.
    }
    \label{table:architecture}
\end{table}

\begin{description}[font=\normalfont\itshape]
    \item[Model Selection (Small model)]
        We use Bayesian Optimization \cite{snoek_practical_2012}
        to find the optimal model. This is done using the Spearmint package \cite{snoek_spearmint_2015}. 
        We ran Bayesian Optimization for one week to find the optimal parameters for the small model. 

    \item[Model Selection (Large model)]
        The large model was found using a combination of grid search and manual tuning. 
        The specific architectures we found is detailed in Table \ref{table:architecture}.
        Bayesian Optimization could not be used because large models were too slow to train.

        After training of the joint model, we also fine-tuned the model by considering each individual task and kickstarting
        the training from the models learned in the joint model. That is, we started training a model whose parameters
        were the same as the multitask model, but the loss function only included one specific task. The loss function for
        task $\tau$, sequence $s$ indexed from $t=1,\ldots,T$ is then 

        $$L(\theta) = -\sum_{s\in S}\sum_{t=1}^T {\ln p_\tau(c_{correct} | f_{t,\tau,s}, \theta)}$$

        This result is labeled as fine-tune in tables \ref{table:architecture} and \ref{table:results}
        We use the validation set during the finetuning to find the best dropout value, but then we include the validation
        set in the retraining set. Dropout generally ensures that early stopping is not needed, so including
        the validation set should improve the accuracy of our model. We fine-tune at a learning rate of $\frac{1}{10}$ of
        the joint model learning rate.

    \item[Time]
        Training of the small model takes 6 hours, while training of the large model takes one day.
        Since testing the fine-tuned models involve passing the data through four separate models,
        while testing the multitask model involves doing all at the same time,
        it takes longer to test on the fine-tuned model. Nevertheless, we were able
        to handle testing speeds of over a million amino acids in under 2 seconds.

    \item[Hardware]
        In order to speed up computation, we utilize the parallel architecture of the GPU,
        which is especially useful for convoultional models which do many parallel computations.
        All training and testing uses a Tesla C2050 GPU unit.
\end{description}

\subsection{Results}

During model selection, we discovered that our model is very robust to model parameters. Most combinations of parameter
tweaks inside the optimal learning rate give a less than 1\% improvement in average accuracy. By using maxpooling with
shift-and-stitch in our model our average accuracy improved by almost 0.5\% with barely any computational slowdown.

Our results on the 4prot dataset are detailed in Table \ref{table:results}.
The small model we found via Bayesian Optimization has approximately as many parameters as previous state-of-the-art
models, but we see that it outperformed the network created by \cite{qi_unified_2012} on all tasks.
Fine-tuning on individual models is necessary for good performance. This implies that it may perhaps be easier
to build an MLP subclassifier for each task, instead of assuming linearity. Training jointly on the large model
already beats \cite{qi_unified_2012}, but finetuning increases the accuracy dramatically.
Additionally, the testing time is reported in milliseconds per million amino acids.
We see that the small models can test fairly quickly,
while the fine-tuned large models have a 2.5$\times$ slowdown. We are the first to report precise training and testing
times for a model on protein property prediction. 

A detailed listing of precision-recall scores for 4prot is given in Table \ref{table:f1}. We see the expected pattern
of lower frequencies having a lower F1 score, since unbalanced datasets are harder to classify.
Precision is very stable, while
recall dramatically lowers according to the frequency of labels. This suggests that our model picked up on several
key properties of labels with few training examples, but missed many. More training data is one way to solve this issue.

Our results on the CullPDB dataset and comparisions with existing state-of-the-art algorithms is detailed in table
\ref{table:resultscullpdb}. We do 1\% better than the previous published best, despite using a dramatically simpler
algorithm. Testing on the CB513 dataset allows a direct comparison to how previous methods perform. We do not achieve a
dramatically higher accuracy rate as we do on 4prot.
We suspect that filtering non-homologuous protein sequences decreases possible accuracy, since we
are essentially demanding a margin of difference between the data distributions for the training and testing samples.
It may not be possible to predict protein properties accurately using a statistical method if non-homologuous
protein sequences were filtered from the training set.

\begin{table}
    \centering
    \begin{tabular}{l | l | l l | l l}
        Task    &   \citeauthor{qi_unified_2012}&  Conv    & fine-     & Conv      & fine-\\
                &                               &   small  &   tuned   &  large    & tuned  \\\hline
        dssp (8)&   68.2                        &  67.0    &   70.6    &   69.5    & \textbf{76.7}    \\
        ssp (3) &   81.7                        &  80.6    &   84.0    &   82.5    & \textbf{89.6}    \\
        sar (2) &   81.1                        &  79.0    &   81.2    &   80.2    & \textbf{84.9}    \\
        saa (2) &   82.6                        &  80.9    &   82.9    &   82.0    & \textbf{86.1}    \\
        Test time& 596k* & 379 & 587 & 553 & 1597
    \end{tabular}
    \caption{
        Q$_c$ accuracy on different architectures of model on 4prot dataset. The number in parenthesis 
        behind the task determines $c$, the number of classes in each task. Testing time is given for
        all tasks simultaneously in \emph{milliseconds per million amino acids}.
        (*) Test time was not detailed in referenced paper, so their
        algorithm was implemented and tested on the CPU.
    }
    \label{table:results}
\end{table}

\begin{table}
    \footnotesize
    \centering
    \begin{tabular}{r | c c c c}
        Per-task Label       &   Recall  &   Precision   &   F1  &   Frequency \\ \hline
        \textbf{dssp} & & & & \\
        H           &    .967   &     .878      &  .920 &    .328       \\
        E           &    .924   &     .821      &  .869 &    .206       \\
        L           &    .748   &     .645      &  .693 &    .211       \\
        T           &    .564   &     .623      &  .592 &    .113       \\
        S           &    .254   &     .621      &  .360 &    .095       \\
        G           &    .363   &     .655      &  .467 &    .035       \\
        B           &    .049   &     .797      &  .093 &    .012       \\
        I           &     0     &      0        &   0   &    .0002      \\
        \textbf{ssp} & & & & \\
        C           &   .875    &   .881        &  .878 &   .418\\
        H           &   .936    &   .919        &  .928 &   .364\\
        E           &   .868    &   .884        &  .876 &   .218\\
        \textbf{sar} & & & & \\
        Inaccessible&.874 & .838& .856& .512\\
        Accessible  &.823 & .861& .842& .488\\
        \textbf{saa} & & & & \\
        Inaccessible& .901&.888 & .894&.650 \\
        Accessible  & .789&.810 & .799&.350 \\
    \end{tabular}
    \caption{
        Recall, precision, and F1 scores for 4prot dataset. Class I for dssp
        does not occur often enough for our model to learn labelings. 
    }
    \label{table:f1}
\end{table}

\begin{table}[!hbt]
    \centering
    \begin{tabular}{l | c}
        Model       &   Q$_8$ \\\hline
        CNF \cite{wang_protein_2011}       &   .649 \\
        GSN \cite{zhou_deep_2014} &   .664 \\
        LSTM \cite{kaae_sonderby_protein_2014}        &   .674 \\
        MUST-CNN (Ours)                    &   \textbf{.684} \\
    \end{tabular}
    \caption{
        Q$_8$ accuracy training on the CullPDB dataset and testing on CB513. Testing takes around the same time
        as for the 4prot dataset. We use the same architecture as MUST-CNN large, detailed in table
        \ref{table:architecture}.
    }
    \label{table:resultscullpdb}
\end{table}

\section{Discussion}

We have described a multilayer shift-and-stitch convolutional architecture for sequence prediction. We use ideas from the image
classification domain to train a deep convolutional network on per-position sequence labeling.
We are the first to use multilayer shift-and-stitch on protein sequences to generate per-position results. Shift-and-stitch
is a trick to quickly compute convolutional network scores on every single window of a sequence at the same time,
but the fixed window sizes of the convolutional network still remains. Surprisingly, 
we achieve better results than whole sequence-based approaches like the
GSN, LSTM, and CNF models used in previous papers (see Table \ref{table:resultscullpdb}). 
We believe this is because the speed of our model allows us to train models with far higher capacity.
We show that the architecturally simpler
MUST-CNN does as well or better than more complex approaches.

In our experiments, the same network works very well on two different large datasets of protein property prediction,
in which we only
changed the amount of dropout regularization. This suggests that our model is very robust and can produce good
results without much manual tuning once we find a good starting set of hyperparameters. More generally, our technique should work on arbitrary per-position sequence tagging tasks, such as 
part of speech tagging and semantic role labeling.

Additionally, our model can make predictions for a million amino acids in under 2 seconds.
Although the main speed bottleneck of protein property prediction is obtaining the PSI-BLAST features,
the speed of our model can be useful on other sequence
prediction tasks where feature extraction is not the bottleneck.

Future work can incorporate techiques such as the fully convolutional network \cite{long_fully_2014} to further
speed up and reduce the parameter set of the model. Another direction is to continue along the lines of LSTMs and
GSNs and try to better model the long range interactions of the protein sequences.

\bibliographystyle{aaai}
\bibliography{citations}

\end{document}